\documentclass[12pt]{article}

\usepackage{sbc-template}
\usepackage{graphicx,url}
\usepackage[utf8]{inputenc}
\usepackage[english]{babel}

\usepackage{graphicx}
\usepackage{xcolor}

\usepackage{multirow}
\usepackage{adjustbox}

\usepackage{algorithm} 
\usepackage{algpseudocode} 
\algnewcommand\algorithmicforeach{\textbf{for each}}
\algdef{S}[FOR]{ForEach}[1]{\algorithmicforeach\ #1\ \algorithmicdo}
     
\title{\texttt{DEBACER}: a method for slicing moderated debates}

\author{Thomas Palmeira Ferraz\inst{1}, Alexandre Alcoforado\inst{1}, Enzo Bustos\inst{1}, \\
André Seidel Oliveira\inst{1}, Rodrigo Gerber\inst{1}, Naíde Müller\inst{3}, \\ 
André Corrêa d'Almeida\inst{4}, Bruno Miguel Veloso\inst{2}, Anna Helena Reali Costa\inst{1}}
\address{Escola Politécnica, Universidade de São Paulo (USP), São Paulo, SP, Brazil
\nextinstitute
Universidade Portucalense \& INESC TEC, Porto, Portugal
\nextinstitute
Catholic University of Portugal, Lisboa, Portugal
\nextinstitute
Columbia University, New York, NY, USA
\email{\{thomas.ferraz,alexandre.alcoforado,enzobustos\}@usp.br,} 
\email{\{rodrigo.gerber,andre.seidel,anna.reali\}@usp.br,}
\email{bruno.m.veloso@inesctec.pt,
  ac3133@sipa.columbia.edu}
}

\begin{document} 

\maketitle

\begin{abstract}
Subjects frequently change in moderated debates with several participants,
such as in parliamentary sessions, electoral debates, and trials.
Partitioning a debate into blocks with the same subject is essential for
understanding. Often a moderator is responsible for defining when a new
block begins so that the task of automatically partitioning a moderated
debate can focus solely on the moderator's behavior. 
In this paper, we (i) propose a new algorithm, \texttt{DEBACER}, which partitions moderated debates; (ii) carry out a comparative study between conventional and BERTimbau pipelines; and (iii) validate \texttt{DEBACER} applying it to the minutes of the Assembly of the Republic of Portugal. Our results show the effectiveness of \texttt{DEBACER}.

% Democracies worldwide produce much textual documentation. The majority of those documents are extensive and challenging to interpret by an ordinary citizen, and much of their content share a standard structure: it is composed of an organized dialogue with a moderator. This happens in Parliament sessions and in trials, to cite two examples. In this paper, we propose the \texttt{DEBACER} algorithm to slice dialogue speeches, dividing them into blocks that share a common subject of discussion. These speech blocks may be used as input for tasks such as topic modeling or sentiment analysis, allowing the data to be presented in a meaningful way to the general public. The main contributions of this paper are a system that calculates statistics from a database of meeting minutes; a new algorithm, \texttt{DEBACER}, which separates blocks of speech from debates; a comparative study between conventional pipelines and BERTimbau for the \texttt{DEBACER} task; and a case study on direct and indirect citations of "corruption" in the minutes of the Portuguese Parliament. Preliminary results show that conventional pipelines can be as effective as BERTimbau, with better efficiency.

\end{abstract}

%\keywords{Natural Language Processing \and Political Documents \and Spoken Text Processing \and speech Split \and Dialogue Partitioning}

%\begin{resumo} 
 % Este meta-artigo descreve o estilo a ser usado na confecção de artigos e
 % resumos de artigos para publicação nos anais das conferências organizadas
  %pela SBC. É solicitada a escrita de resumo e abstract apenas para os artigos
  %escritos em português. Artigos em inglês deverão apresentar apenas abstract.
  %Nos dois casos, o autor deve tomar cuidado para que o resumo (e o abstract)
  %não ultrapassem 10 linhas cada, sendo que ambos devem estar na primeira
  %página do artigo.
%\end{resumo}

\section{Introduction}
\label{sec:introduction}

During a long debate with numerous participants, it is usual to have several subject changes. Partitioning these dialogues into \emph{blocks of speeches} that represent different stages of the debate is essential to understand what is being said by each participant and to perform various natural language processing tasks. This partitioning of dialogues can be beneficial when one wants to analyze and evaluate the debate, \textit{e.g.}, tracking the evolution of an organization's meetings over several years, indicating how many times a member of the organization has participated in discussions about a particular subject of interest. Thus, this dialogue partitioning procedure can benefit sentiment analysis, position-taking detection, argument agreements, and topic modeling tasks.

Current dialogue state tracking approaches model the entire dialogue as an input graph \cite{ales2018extraction}, or even model the dialogue as a sequential problem, using techniques such as recurrent neural networks \cite{wen2017network} and attention models \cite{shan2020contextual} to capture the state of the dialog. While these methods can be very efficient in real-time applications that assist other agents in taking actions, such as in conversational AI, they do not address the existence of multiple participants in the conversation. Besides, they are too complex for just partitioning the dialogues into blocks of speech, especially when the characteristics of those dialogues can guide such partitioning.

This is the case of \emph{moderated debates}, forums in which the discussion is organized and controlled by a person in charge of giving the floor to each of the participants in turn. An excellent example of moderated debate is a parliamentary session, where the representatives participate in public hearings and formulate new laws. The documents produced by these discussions are extensive and complex for an ordinary citizen to understand and search for specific information. Electoral debates, public hearings, and trials are other examples of dialogues organized with a moderator. We here argue that the moderator plays an essential role in these discussions, defining the maintenance of the current block of speeches or the beginning of a new one. In this case, it is possible to transform this complex problem into a simpler one that does not require processing the entire history of the dialogue or even the use of complex representations such as graphs.

In this work, we contribute a new algorithm, \texttt{DEBACER}, that uses a machine-learning classification pipeline to slice moderated debates, helping to extract insightful information from their transcripts. We validate this method in a case of moderated debates in Portuguese: the Assembly of the Republic of Portugal. We assess the \emph{political statements}, the opening moments of sessions when participants can bring new topics or respond to the previous speaker. In these moments, the chairperson acts as a moderator, managing the entire dialogue: interrupts a very long speech, allows retorts or complements to a statement, or even gives the floor to another interlocutor on a new subject.
We also make a comparative study of what is the best classification pipeline for this task. One of the pipelines is BERTimbau \cite{souza2020bertimbau}, the Portuguese version of BERT that uses a neural architecture based on attention models and has been reported as the state-of-the-art for several text classification tasks \cite{devlin2019bert}. We fine-tune BERTimbau for our problem and compare it to conventional text classification pipelines. We also seek to define which aspects are domain-dependent and domain-independent when we apply the proposed method.

Therefore, the contributions of this paper are:
\begin{enumerate}
  \item A new algorithm, \texttt{DEBACER}, which separates speech blocks from debating with a moderator;
  %which separates the blocks of speech of debates with a moderator; 
  \item A comparative study among conventional pipelines and BERTimbau for the \texttt{DEBACER} task, defining the best of them and evaluating domain dependencies;% and %defining the best pipeline and evaluating domain dependence aspects; and
  \item The application of \texttt{DEBACER} in a set of minutes of the Portuguese Parliament.
\end{enumerate}

\section{Related Work}
\label{sec:relatedwork}
%político / computacional / mais específico em diálogos com moderador.
%Text Mining for democracy
This section presents some relevant work published that uses machine learning and natural language processing techniques in debates with a moderator.

Guerini, Strapparava, and Stock \cite{guerini2008corps} propose the tagging of political speeches with audience reactions for further automatic analysis. The reaction acts as a validation of the rhetoric of a political party. The authors search for a set of keywords identifying the audience's state and then apply the TextPro and SentiWordNet to compute the persuasive impact of the speech. In our work, we want to identify the speech transition among several politicians automatically. In the minutes, there are also some reactions that we will use for further research on persuasive speech.

Yu, Kaufmann, and Diermeier \cite{yu2008classifying} developed a framework for classifying party affiliation from political speeches. The authors trained Naive Bayes and SVM classifiers using the 2005House dataset to validate the Senate speeches. The authors found that the speeches contain a time-dependency pattern and more recent data drives thru a better classification. Our work differs on identifying individuals and not political parties. In terms of techniques, we also employ a more sophisticated classification pipeline using BERTimbau.

Lippi and Torroni \cite{lippi2016argument} presents an automatic extraction algorithm to capture arguments and claims from UK politicians. The pipeline is composed of three modules speech recognition system, feature extraction (Bag-of-Words, part-of-speech tags, and lemmas), and a classifier (SVM). Our proposed pipeline is similar to this work, but we aim to identify the speaker in a specific part of the speech. We use more robust classifiers that can compete with state-of-the-art models (BERTimbau).

Roush and Balaji \cite{roush2020debatesum} proposes a model called debate2vec, which used a trained model using a dataset containing text from public debates on the Parliament. The model uses a set of fast text word vectors previously described by Bojanowski \textit{et al.} \cite{bojanowski2017enriching}. The focus of the model is to classify arguments on political speeches correctly. 
Our work differs in identifying individuals rather than political arguments.
%Our results show that with a lighter model, we can obtain results that are at least competitive compared to BERTimbau, which is a very current architecture, but costly in computational and data terms.

\section{Proposal}
\label{sec:proposedframework}

% \subsection{Task Formalization}
% \label{subsec:taskformalization}
Speeches given during a meeting follow a particular chronology and are transcript into minutes.  %, i.e., they are written in minutes. 
Each \textit{speech} $\mathcal{S}$ has a \textit{debater} $\mathcal{D}$ who has the floor, and its content is composed of a sequence of $n$ uttered \textit{words} $\mathcal{S_D} = (w_{1}, w_{2}, w_{3}, \ldots, w_{n})$. 
%(\mathcal{W}_{1}, \mathcal{W}_{2}, \mathcal{W}_{3}, \ldots, \mathcal{W}_{n})$. 
We denote an \textit{agenda item} $\mathcal{A}$, as a sequence of $m$ speeches $\mathcal{A} = (\mathcal{S}_{1}, \mathcal{S}_{2}, \mathcal{S}_{3}, \ldots, \mathcal{S}_{m})$ related to the same meeting item. 
A \textit{minute} $\mathcal{M}$ is the sequence of all agenda items,  $\mathcal{M} = (\mathcal{A}_{1}, \mathcal{A}_{2}, \ldots ,\mathcal{A}_{l})$ that make up a given meeting, having a unique identification for each meeting.

It is usual for a person's speech to either be a statement about something discussed in the immediately preceding speech or introducing a new subject. Our objective is to partition the speeches in the political statements to identify a sequence of speeches that refer to the same subject. 
To do so, we define a \textit{speech block} $\mathcal{B}_{ij} = (\mathcal{S}_{i}, \mathcal{S}_{i+1},\ldots, \mathcal{S}_{j})$ as a subsequence of $\mathcal{A}$, such that $\forall k \in (i,j]$, $\mathcal{S}_{k}$
is a speech that follows $\mathcal{S}_{i}$ logically, i.e., a block refers to speeches about the same subject. 
The purpose of the algorithm proposed here, \texttt{DEBACER}, is precisely to partition each agenda item $\mathcal{A}_{x}$ into a sequence of blocks $\mathcal{B}_{ij}^{x}$, such that
$\mathcal{B}_{1i}^{x} \cap \mathcal{B}_{ij}^{x} \ldots \cap \mathcal{B}_{mn}^{x} = \emptyset$ and $\mathcal{B}_{1i}^{x} \cup \mathcal{B}_{ij}^{x} \ldots \cup \mathcal{B}_{mn}^{x} = \mathcal{A}_{x}$.

\subsection{The \texttt{DEBACER} Algorithm}
\label{subsec:debacer}

% \textcolor{red}{\texttt{DEBACER} can be applied to domains of debates with a moderator, a person who leads the debate and controls speeches, establishing the next debater to speak and maintaining or changing the subject of the discussion. 
% Thus, the partition of the speeches can be strongly guided by the moderator's speech.
% This is precisely the case for the sessions of the Portuguese Parliament.
% Political statements are the agenda item where congress members can bring important issues to Parliament.
% By taking the floor, the member can either present a new subject or comment on a subject that is already under discussion. The chairperson of the session acts as moderator and, most of the time makes it clear where the subsequent intervention fits.}

%Initially, \texttt{DEBACER} separates from $\Pi_{\mathcal{M}}$ the parts of the minutes referring to political statements, placing them in $\Phi_{\mathcal A_{PS}}$ (steps 2 to 9 of Algorithm \ref{alg:\texttt{DEBACER}}). 

From a set of minutes, $\Pi_{\mathcal{M}}$ containing transcripts of moderated debates, \texttt{DEBACER} divides all agenda items into blocks of speech. For each agenda item, $\mathcal A$ inside each minute $\mathcal{M}$, \texttt{DEBACER} runs through all moderator's speeches and identifies if their content indicates an interruption in the subject.
Two functions are used (step 8 in Algorithm \ref{alg:debacer}): 
\begin{description}
    \item \textsc{IsModerator}: Check if the debater is the moderator. It returns TRUE when the debater of $\mathcal{S}$ is the current moderator (\textit{i.e.}, the current chairperson). This information is extracted from the database, where there is a special marker to indicate who the moderator is.
    \item \textsc{IsSubjectInterruption}: Check if the content is classified as an interruption in the subject. It uses a domain-dependent text classification pipeline $\mathcal{C}$ to find out whether the moderator's speech content matches an interrupt or not.
\end{description}
A new speech block starts if the moderator changes the subject (steps 9 to 12). Otherwise, the speech belongs to the current speech block (step 13).
At the end of the process, the database contains labels from the speech block each speech of the political declarations belongs, of each of the minutes in $\Pi_{\mathcal{M}}$ (step 17).

%Function \textsc{IsModerator} .%, which is expected to be similar to the  in Figure \ref{fig:Transcripts Database}, when the name of the debater is equal to \textit{``Presidente"} (chairperson in Portuguese). 
%It is indicated in the parliamentary minutes and stored in the database during the offline phase.

%Function \textsc{IsSubjectInterruption} 

\begin{algorithm}[htb]
    \footnotesize
	\caption{Split the political statements from the minutes into speech blocks} 
	\label{alg:debacer}
    \textbf{Inputs:} a set of minutes $\Pi_{\mathcal{M}}$ (the database), and a subject interruption classifier $\mathcal{C}$%\\
    %\textbf{Output:} 
    %{ \fontsize{7}{7}\selectfont
	\begin{algorithmic}[1]
    	\Procedure{DEBACER}{$\Pi_{\mathcal{M}},\mathcal{C}$} 
    	%\State $\Phi_{\mathcal A_{PS}} \leftarrow \{\} $ \Comment{Set of agenda items to be splited}
    	\ForEach {$\mathcal M \in \Pi_{\mathcal{M}} $}
    	    \ForEach {$\mathcal A \in \mathcal M$}
        	    %\If{\textsc{IsPoliticalStatement}($\mathcal A$)}
        	        %\State $\Phi_{\mathcal A_{PS}} \leftarrow \Phi_{\mathcal A_{PS}} \cup \{ \mathcal A \} $ \Comment{Selects items of interest}
        	    %\EndIf
    	    %\EndFor
    %	\EndFor
    	%\ForEach {$\mathcal A \in \Phi_{\mathcal A_{PS}} $}
    	    \State $\mathcal A.blocks \leftarrow \{\}$
    	    \State $i \leftarrow 0$
    	    \State $\mathcal B_{i} \leftarrow \{\}$
    	    \ForEach {$s \in \mathcal A$}
        	    \If{\textsc{IsModerator}($s.debater$)$\wedge$  \textsc{IsSubjectInterruption}($\mathcal{C},s.content$)}
        	        \State $\mathcal A.blocks \leftarrow \mathcal A.blocks \cup \mathcal B_{i}$
        	        \State $i \leftarrow i + 1$
        	         \State $\mathcal B_{i} \leftarrow \{\}$
        	    \EndIf
        	     \State $\mathcal B_{i} \leftarrow \mathcal B_{i} \cup \{ s \}$
    	    \EndFor
    	    \State $\mathcal A.blocks \leftarrow \mathcal A.blocks \cup \mathcal B_{i}$
    	\EndFor
    	\State \textsc{Update}($\Pi_{\mathcal{M}}, \mathcal A$) \Comment{Updates the database $\Pi_{\mathcal{M}}$ with $\mathcal A$ sliced}
    	%\State \textbf{return} $\Pi_{\mathcal{M}}$
    	\EndFor
    	\EndProcedure
	\end{algorithmic}%}
\end{algorithm}

\subsection{Domain-dependent Aspects}
\label{subsec:domaindependent_aspects}

%\textcolor{red}{******** Precisa resumir e realinhar daqui até o fim do capítulo **********}
%\\
%----NEW----
%For the DEBACER algorithm to be applied to different data domains, changes should be made to some aspects of the framework; otherwise, accuracy may decrease. 

Although \texttt{DEBACER} is domain-independent and can be applied to problems that match the proposed moderated debate problem description, it requires a properly arranged database and a domain-dependent text classification pipeline to detect interruptions. One premise is that the database in which the \texttt{DEBACER} will be executed is composed of 1 column with textual data (the speeches $(\mathcal{S}_{1}, \mathcal{S}_{2}, \mathcal{S}_{3}, \ldots, \mathcal{S}_{m})$ delivered) and 1 column with its corresponding authors (the \textit{debaters} $\mathcal{D}_{i}$). 
Considering a database in this format, it is essential to train the specific classification pipeline for the data domain. We can achieve that by annotating a training dataset from the database and applying it to a pipeline of supervised learning methods. These methods, at the end of the training, should classify ``1" \ for the interruption -- when a speech block $\mathcal{B}_{ij}$ ends and another block $\mathcal{B}_{(j+1)k}$ is initiated -- and ``0" \  otherwise. In the experiments in Section \ref{sec:experimentalsetup}, we evaluate several pipelines and recommend the best one for this, which may involve text cleaning, feature selection, and different classifiers.

A fundamental aspect to be considered when training the classifier is that, in general, the data will be \emph{inherently imbalanced}, that is, an uneven distribution of target groups is a characteristic of the problem. In the case of a debate, in speaker transitions, there is more continuation (``0") than subject change (``1"). Not properly dealing with this problem can lead to classifier bias and poor performance. It is possible to treat the imbalance problem at two levels: data and algorithm. Data-level methods can take into account techniques such as \emph{Stratified $K$-fold Cross-validation} (CV) and a frequency matching of classes applying \emph{Dataset Resampling} or \emph{Data Augmentation}. On the other hand, algorithmic-level methods will take into account the balancing mechanisms of the training algorithms for each type of model (for example, Random Forest applies balanced sub-sampling) and which performance metrics are used to compare them, especially while doing hyperparameters search. In section \ref{sec:experimentalsetup}, we apply some of these methods: a modified version of $K$-fold CV, algorithms that somehow deal with data imbalance (\emph{BERT, LR, SVM, RF}), and smart metrics for this problem: \emph{F1-Score}, \emph{Cross-Entropy} and \emph{Brier Score}.
This new pipeline should then be applied to the database to determine which speeches delivered by the moderator are interruptions and which are not.

\section{Experimental Setup}
\label{sec:experimentalsetup}

%\textcolor{red}{Resumir, focar em dizer que os experimentos servem a validar o método proposto} Our experiments aim to find the best pipeline for the chairperson's speech classifier introduced in Section \ref{subsec:debacer} and, by applying algorithmic-level and data-level imbalance correction methods, to verify whether it is possible to obtain similar or superior results to BERTimbau, which was shown in literature the best results for text classification in Portuguese.
%For this, we compare BERTimbau to four other common features for language processing. 
Our experiments aim to validate the algorithm proposed in Section \ref{subsec:debacer}, as well as to find the best pipeline for the moderator's speech classifier. To this end, we make use of minutes of sessions of the Portuguese Parliament and compare BERTimbau, which has presented the best results in the literature for classification of texts in Portuguese, with four other common features for language processing.

\subsection{Data}

%\subsubsection{Data Collection} 

\begin{figure}[htb]
    \centering
    \includegraphics[width=0.9\textwidth]{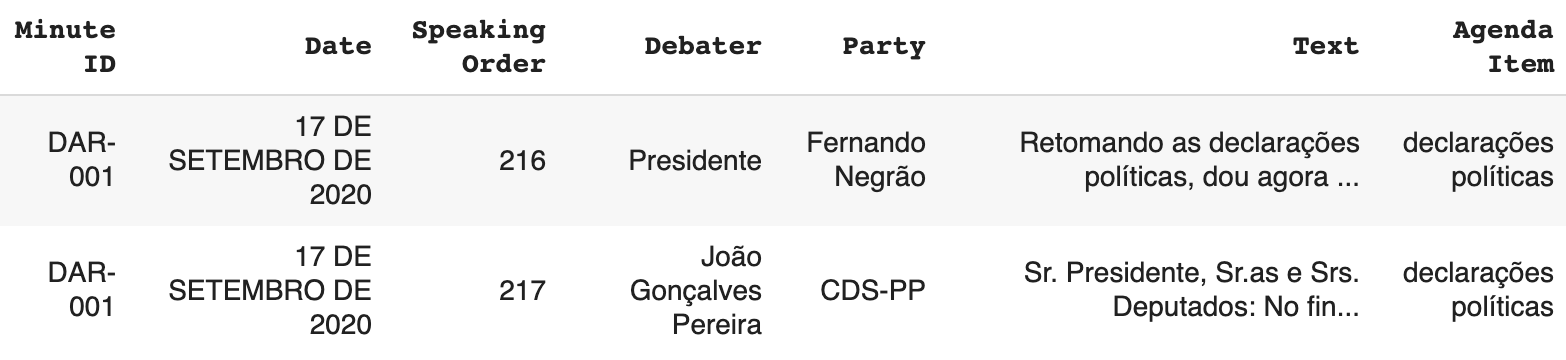}
    \caption{Transcripts Database.}
    \label{fig:Transcripts Database}
\end{figure}

\textbf{Data Collection} We leverage data from the Portuguese Parliament (\textit{Assembleia da República}) website\footnote{https://debates.parlamento.pt/} by using a web crawler algorithm specifically designed for the task of downloading the minutes in TXT format. These minutes were then separated into individual speeches, which were organized into a structured database, consisting of the fields: minute id, date, speaking order, debater, party, text, and agenda item, as illustrated in Figure \ref{fig:Transcripts Database}. The current legislature minutes were used in this work, the XIV Legislature of the Portuguese Republic (from 2020/09/16 to 2021/02/25). Once ready, the database was composed of 20543 rows. For \texttt{DEBACER} application purposes, we selected from this database only the agenda item ``political statements" that correspond to the moment of the parliamentary meeting in which members can openly discuss different topics, introducing new subjects, or commenting on previous ones.

%\subsubsection{Data Annotation} 

\noindent \textbf{Data Annotation} It is necessary to annotate a dataset in order to apply it to a supervised learning pipeline that, at the end of the training, classifies ``1" for the interruption and ``0" otherwise. In the case of the Portuguese Parliament, the annotation process was carried out in a semi-automatic mode. We sampled and manually labeled about 70 speeches pronounced by chairpeople (\textit{i.e.}, the moderator). This annotated data was used to train a Random Forest classifier which was used to label the remaining unlabeled data. Then, a manual review was carried out, observing each chairperson's speech and its context (speeches of previous and subsequent debaters) to correct misclassifications and revalidate the premise that we could reduce the partitioning speeches only based on the chairperson's speeches. The labeling process with the Random Forest classifier considerably reduced the human efforts involved in the annotation. Table \ref{tab:presiding} presents the distributions of the resulting annotated dataset.

% Please add the following required packages to your document preamble:
% Please add the following required packages to your document preamble:
% Please add the following required packages to your document preamble:
% Please add the following required packages to your document preamble:
% \usepackage{multirow}
% Please add the following required packages to your document preamble:
% \usepackage{multirow}
\begin{table}[]
\centering
\caption{Target variable distribution. 1 is for subject interrupt and 0 otherwise.}
\label{tab:presiding}
\begin{adjustbox}{width=0.7\textwidth,center}
\begin{tabular}{|c||ccccc||c|}
\hline
\multirow{2}{*}{\textbf{\begin{tabular}[c]{@{}c@{}}Target\\  Variable\end{tabular}}} & \multicolumn{5}{c||}{\textbf{Moderator}}                                                                                                                                                                                                                                                                                                                 & \multirow{2}{*}{\textbf{Total}} \\ \cline{2-6} 
                                                                                     & \textbf{\begin{tabular}[c]{@{}c@{}}José M.\\ Pureza\end{tabular}} & \textbf{\begin{tabular}[c]{@{}c@{}}Eduardo F.\\ Rodrigues\end{tabular}} & \textbf{\begin{tabular}[c]{@{}c@{}}Edite\\ Estrela\end{tabular}} & \textbf{\begin{tabular}[c]{@{}c@{}}António\\ Filipe\end{tabular}} & \textbf{\begin{tabular}[c]{@{}c@{}}Fernando\\ Negrão\end{tabular}} &                                 \\ \hline
\multirow{2}{*}{\textbf{\#0}}                                                        & \multirow{2}{*}{165}                                              & \multirow{2}{*}{147}                                                    & \multirow{2}{*}{99}                                              & \multirow{2}{*}{69}                                               & \multirow{2}{*}{69}                                                & \multirow{2}{*}{\textbf{549}}   \\
                                                                                     &                                                                   &                                                                         &                                                                  &                                                                   &                                                                    &                                 \\ \hline
\multirow{2}{*}{\textbf{\#1}}                                                        & \multirow{2}{*}{10}                                               & \multirow{2}{*}{14}                                                     & \multirow{2}{*}{5}                                               & \multirow{2}{*}{7}                                                & \multirow{2}{*}{5}                                                 & \multirow{2}{*}{\textbf{41}}    \\
                                                                                     &                                                                   &                                                                         &                                                                  &                                                                   &                                                                    &                                 \\ \hline
\multirow{2}{*}{\textbf{Total}}                                                      & \multirow{2}{*}{\textbf{175}}                                     & \multirow{2}{*}{\textbf{161}}                                           & \multirow{2}{*}{\textbf{104}}                                    & \multirow{2}{*}{\textbf{76}}                                      & \multirow{2}{*}{\textbf{74}}                                       & \multirow{2}{*}{\textbf{590}}   \\
                                                                                     &                                                                   &                                                                         &                                                                  &                                                                   &                                                                    &                                 \\ \hline
\end{tabular}
\end{adjustbox} 

\end{table}

%\subsubsection{Multi-label Stratified K-fold} 
%\label{subsec:mlskfold}

\noindent \textbf{Multi-label Stratified $K$-fold}
Cross-validation is a widely used technique to assess the generalization of a model. In particular, the $K$-fold is a well-known cross-validation method that consists of randomly dividing the dataset into $K$ non-intersecting (mutually exclusive) sets, then training the model $K$ times, covering all possible combinations of having the union of $K-1$ folds as the training set and the remaining fold as the test set. However, some problems present us with situations in which the uniformity of these $K$ folds may be impaired: it may happen because the problem has more than one target variable or, in the Portuguese Parliament case, because factor others than the target variable influence the data distribution. We specifically mention two factors that may affect the performance of the classifier: (i) the debater, because each person has, within its individuality, its vocabulary preference; and (ii) the time, language is dynamic, and overtime terms become outdated, and new ones appear throughout years of parliamentary sessions. \cite{sechidis2011stratification} proposes a stratified $K$-fold for multiple variables to have a population distribution in the subgroups more faithful to the parent group. Considering the short time interval between the minutes processed, we chose not to consider the time variable in this problem. Instead, \textbf{we employ the Multi-label Stratified $K$-fold approach, only taking the debater variable ($\mathcal{D}^{*}$) and the target variable (0 or 1) as labels}.

\subsection{Baselines}
\begin{figure}[htb]
    \centering
    \includegraphics[width=0.95\textwidth]{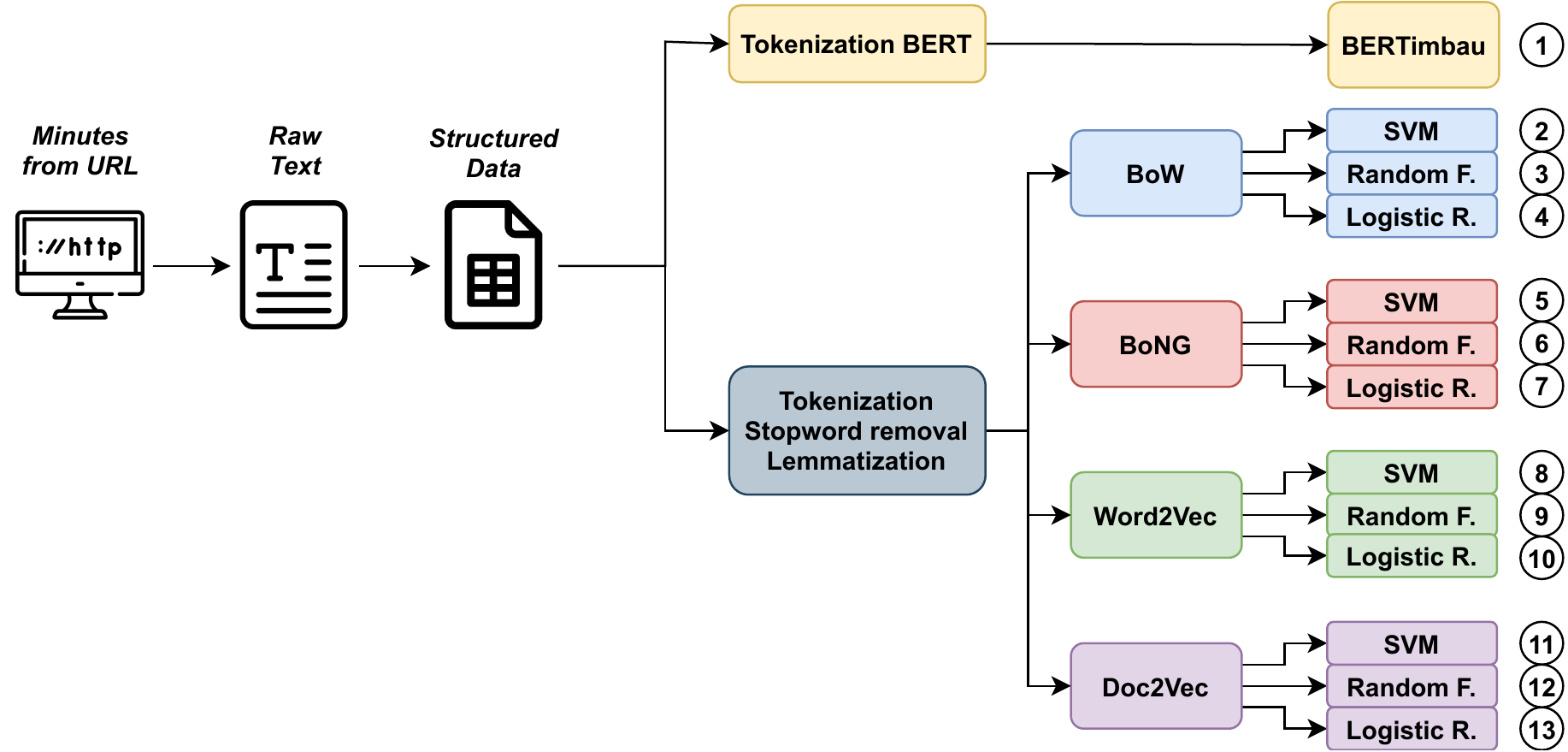}
    \caption{Scheme representing the pipelines configurations to validate \texttt{DEBACER}.}
    \label{fig:flowchart}
\end{figure}

%We consider the following baseline methods:
\begin{description}
\item [BERTimbau:] The BERT architecture consists of 12 Transformers blocks, each block has a hidden size of 768 and 12 self-attention heads. We fine-tuned BERTimbau pre-trained model, adding three dense layers (64 ReLU-32 ReLU - 1 Sigmoid), with a dropout of 0.2 between them, with all BERT layers unfrozen (totalizing about 334M training parameters), a learning rate of $10^{-5}$, AdamW optimizer, and Binary Cross-Entropy as the loss function.
\item [Bag-of-Words (BoW):] the most straightforward text feature. It is a sparse vector of the frequency of words in a text whose size is equal to the vocabulary size.
\item [Bag-of-N-Grams (BoNG):] a derivation of BoW, a vector of frequencies of the $N$-Grams present in the text, \emph{i.e.}, the count of all possible appearances of specific $N$ words in a row. We use $N=3$, \emph{i.e.}, unigrams, bigrams, and trigrams were used in our frequency vector. The size of this vector can be large, so we use feature selectors to choose which $N$-Grams are most relevant. Some classification algorithms already have this built-in (like Random Forest), but when it is not, we employ \emph{Truncated SVD} \cite{kim2005dimension}, which applies Single Value Decomposition for dimensionality reduction in sparse matrices. So the BoNG acts as a frequency vector of relevant expressions.
\item [Word2Vec:] in this configuration, we train a \emph{Word Embedding} representation of each word on  the  entire  base  of  Portuguese  Parliament  minutes  (from 2020/09/16 to 2021/02/25), using \emph{word2vec} method \cite{mikolov2013efficient}. We use size $n = 100$ and take the average of the vectors to provide a \emph{Sentence Embedding} representation of each data. 
\item [Doc2Vec:] we train a \emph{Sentence Embedding} representation of each data, also on the entire base, using the \emph{doc2vec} method \cite{le2014distributed} with $n = 50$.
\end{description}
For \textbf{BoW}, \textbf{BoNG}, \textbf{Word2Vec} and \textbf{Doc2Vec} configurations, the data is pre-processed before being used for training. We employ  \emph{Tokenization} (segmenting a text into small significant units), \emph{Stopword Removal} (cutting non-significant parts of the vocabulary such as articles, connectives, prepositions) and \emph{Lemmatization} (converting nouns and adjectives to their masculine and singular form and transforming the existing verbs into their infinitive form in order to reduce vocabulary size and promote the abstraction of the word meaning). For these configurations, at the end of the pipelines, we apply different classifiers that have been reported to perform well for imbalanced data text classification: \textbf{Logistic Regression (LR)} \cite{fernandez2018learning}, \textbf{SVM} \cite{liu2009imbalanced} and \textbf{Random Forest (Random F.)}
\cite{wu2014forestexter}. Figure \ref{fig:flowchart} outlines the pipeline configurations to be evaluated in this experiment.

\subsection{Performance Metrics}
Considering the inherently imbalanced nature of the data in this task, 
%as described in section \ref{subsec:domaindependent_aspects}, 
we chose metrics that help offset this problem and compare models more fairly. The following three metrics are used, in order of importance:

\begin{description}
\item \textbf{F1-score:} was evaluating based on accuracy tends to disregard the minority class. This can be solved using $Recall = \frac{TP}{TP+FN}$. However, using only recall makes us lose control of the prediction quality, so we define $Precision = \frac{TP}{TP+FP}$. The F1-score is the harmonic mean between both. So, when we maximize the F1-score, we maximize the gains in both properties. 
\item \textbf{Cross-entropy:} quantifies the average difference between predicted and expected probability distributions, being defined as $CE = - \frac{1}{N} \Sigma_{i}(1 - y_{i}) log(1 - f(x_{i})) + y_{i} log(f(x_{i}))$, where $y_{i}$ is the target probability and $f(x_{i})$ is the learned function.
\item \textbf{Brier Score:} measures the certainty the classifier has in its predictions, computing the mean squared error (MSE) between predicted probability scores and the true class indicator, where the positive class is coded as ``1", and negative class as ``0" ~\cite{fernandez2018learning}. We use a modified version of this metric, proposed by \cite{wallace2012class}, that consists of decomposing this metric for each class, which for the positive would be defined as $BS^{+} = \frac{1}{N_{y_{i}=1}} \Sigma_{y_{i}=1} (y_{i}-f(x_{i}))^{2}$.
\end{description}
Wilcoxon-Holm post-hoc analysis \cite{IsmailFawaz2018deep} is used to compare the performance of the pipelines used against BERTimbau. 
% This makes use of the Wilcoxon test \cite{wilcoxon1992individual}, a non-parametric statistical hypothesis test to detect pairwise difference significance and rank populational means, which is an alternative to the traditional Student's t-test as it does not require a normal distribution. 
The test significance is $\alpha = 0.05$.

\subsection{Implementation Details}
The \texttt{DEBACER} algorithm and the Portuguese parliament minutes database were implemented using the Pandas library. We implemented BERTimbau from Hugging Face using the TensorFlow framework. Other baselines made use of the Scikit-learn library.  We performed hyperparameter optimization by Bayesian search using Tune framework\footnote{https://docs.ray.io/en/master/tune/}. \textbf{We use the F1-score metric for model ranking and selection}. Winning models for each configuration are shown in Table \ref{tab:hyperparameters}. The experiments were run on an Intel Xeon 2-core 2.30~GHz CPU and an Nvidia T4 16~GB 1.59~GHz GPU. Table \ref{tab:hyperparameters} shows the winner hyperparameters for each pipeline configuration. %\textcolor{red}{devemos falar que LR e Random f tem parametros de class weight for imbalance?? O algortihmo \texttt{DEBACER} foi implementado usando python e pandas (para anotar e update a database)}

\begin{table}[htb]
\caption{The winning hyperparameters for each pipeline configuration}
\label{tab:hyperparameters}
\centering
\begin{adjustbox}{width=0.65\textwidth,center}
\begin{tabular}{|c|c|c|}
\hline
Features                  & Best Classifier & Hyperparameters                                                                                                      \\ \hline
\multirow{3}{*}{BoW}      & SVM             & C=482.2 kernel=`linear'                                                                                              \\ \cline{2-3} 
                          & Random F.       & \begin{tabular}[c]{@{}c@{}}criterion=`gini' \\ class\_weight=`balanced\_subsample' \\ n\_estimators=357\end{tabular} \\ \cline{2-3} 
                          & LR              & \begin{tabular}[c]{@{}c@{}}solver=`LBFGS' penalty=`L2'\\  class\_weight=None C=1.02\end{tabular}                     \\ \hline
\multirow{3}{*}{BoNG}     & SVM             & \begin{tabular}[c]{@{}c@{}}n\_TSVD=188 kernel=`linear' \\ C=7932\end{tabular}                                        \\ \cline{2-3} 
                          & Random F.       & \begin{tabular}[c]{@{}c@{}}criterion=`gini' class\_weight=None \\ n\_estimators=680\end{tabular}                     \\ \cline{2-3} 
                          & LR              & \begin{tabular}[c]{@{}c@{}}n\_TSVD=148 solver=`SAGA' \\ penalty=`L1' class\_weight=None C=20.1\end{tabular}          \\ \hline
\multirow{3}{*}{word2vec} & SVM             & \begin{tabular}[c]{@{}c@{}}kernel=`rbf' C=3.75\end{tabular}                                                       \\ \cline{2-3} 
                          & Random F.       & \begin{tabular}[c]{@{}c@{}}criterion=`entropy' class\_weight=None \\ n\_estimators=238\end{tabular}                     \\ \cline{2-3} 
                          & LR              & \begin{tabular}[c]{@{}c@{}}solver=`saga' penalty=`L1' \\ class\_weight=None C=37.05\end{tabular}                     \\ \hline
\multirow{3}{*}{doc2vec}  & SVM             & kernel=`linear' C=176.36                                                                                             \\ \cline{2-3} 
                          & Random F.       & \begin{tabular}[c]{@{}c@{}}criterion=`gini' class\_weight=None \\ n\_estimators=231\end{tabular}                     \\ \cline{2-3} 
                          & LR              & \begin{tabular}[c]{@{}c@{}}solver=`LBFGS' penalty=`L2' \\ class\_weight=`balanced' C=97448\end{tabular}              \\ \hline
\end{tabular}
\end{adjustbox}
\end{table}

\section{Results and Discussion}
\label{sec:resultsdiscussion}

%Table \ref{tab: results} shows the results of the pipelines compared for detecting subject change.
%Figure \ref{fig: diagram} shows the critical difference diagram, which graphically presents the result of the Wilcoxon-Holm post-hoc analysis by pairwise statistical difference comparison. 
%Results in table \ref{tab: results} show some differences between the methods, mainly for the f1-score metric, which is our focus of analysis. However, the post-hoc analysis allows us to state that there is no statistical difference between any of the tested pipelines. This means that it is possible to build a simpler classification pipeline (such as the one based on Bag-of-Words) that has a performance comparable to the state-of-the-art BERTimbau.

\begin{figure}[htb]
    \centering
    \includegraphics[width=0.8\textwidth]{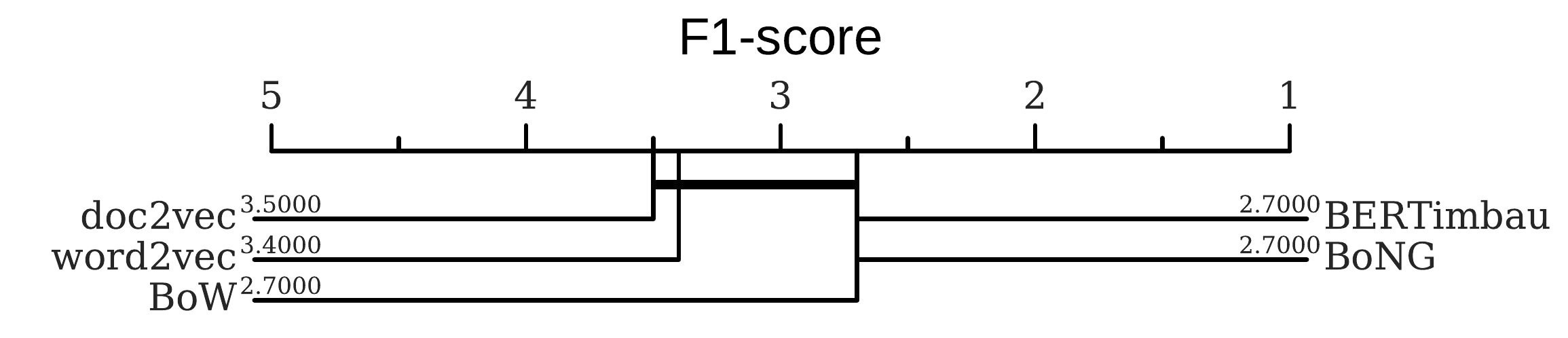}
    \caption{Critical difference diagram showing pairwise statistical difference comparison of the five pipeline configurations for detecting subject change.}
    \label{fig:diagram}
\end{figure}

Table \ref{tab:results} shows the results of the pipelines compared for detecting subject change. When comparing the proposed baselines, it is remarkable that BERTimbau immediately presents an excellent performance: it shows a good F1-Score ($97.5~\%$), the lowest Cross-Entropy ($0.010$) and one of the lowest Brier Score Positive ($0.025$). Smaller Cross-Entropy means that the model generalized better the problem as a whole. The lower the Brier Score Positive, the more confident the classifier is in its predictions about the target class ($ ``1" $). However, the best comparison metric is the F1-Score, and in this configuration, the sparse representation by Bag-of-N-Grams performed numerically above BERTimbau (with $97.8~\%$), while the sparse representation by Bag-of-Words tied with BERTimbau. The continuous representations (doc2vec and word2vec) performed numerically slightly below the others. Nevertheless, Wilcoxon-Holm post-hoc analysis by pairwise statistical difference comparison presents a statistical tie between the best versions achieved for each pipeline. The critical difference diagram is in Figure \ref{fig:diagram}.

\begin{table}[ht]
\caption{Experimental results}
\label{tab:results}
\centering
\begin{adjustbox}{width=0.9\textwidth,center}
\begin{tabular}{|c|c|c|c|c|c|}
\hline
\multirow{2}{*}{Features} & \multicolumn{1}{c|}{\multirow{2}{*}{Classifier}} & \multicolumn{1}{c|}{\multirow{2}{*}{F1-score}} & \multicolumn{1}{c|}{\multirow{2}{*}{Cross-Entropy}} & \multirow{2}{*}{Brier-Score +} & \multirow{2}{*}{Time} \\
                          & \multicolumn{1}{c|}{}                                 & \multicolumn{1}{c|}{}                          & \multicolumn{1}{c|}{}                               &                                        &                       \\ \hline \hline
BERT Tokens               & \multicolumn{1}{c|}{BERTimbau}                        & \multicolumn{1}{c|}{$0.975 \pm 0.031$}         & \multicolumn{1}{c|}{$0.010 \pm 0.007$}               & $0.025 \pm 0.034$                      & $4min40s$               \\ \hline
\multirow{3}{*}{BoW}      & SVM                                                   & $0.968 \pm 0.063$                              & $0.019 \pm 0.016$                                   & $0.025 \pm 0.020$                      & $0.1 s $                \\ 
                          & Random F.                                             & $0.975 \pm 0.031$                              & $0.036 \pm 0.007$                                   & $0.067 \pm 0.027$                      & $3.1 s$                 \\  
                          & LR                                                    & $0.975 \pm 0.031$                              & $0.025 \pm 0.012$                                   & $0.038 \pm 0.016$                      & $0.1 s$                 \\ \hline
\multirow{3}{*}{\textbf{BoNG}}     & SVM                                                   & $0.978 \pm 0.044$                              & $0.020 \pm 0.010$                                     & $0.041 \pm 0.020$                       & $1.4 s$                \\ 
                          & Random F.                                             & $0.976 \pm 0.029$                              & $0.041 \pm 0.005$                                   & $0.103 \pm 0.011$                      & $5.1 s$                 \\ 
                          & \textbf{LR}                                                    & $\textbf{0.978} \pm \textbf{0.044}$                              & $\textbf{0.018} \pm \textbf{0.009}$                                   & $\textbf{0.037} \pm \textbf{0.021}$                      & $\textbf{3.8 s}$                 \\ \hline
\multirow{3}{*}{word2vec} & SVM                                                   & $0.936 \pm 0.058$                              & $0.028 \pm 0.010$                                    & $0.067 \pm 0.074$                      & $0.2 s$                 \\ 
                          & Random F.                                             & $0.889 \pm 0.055$                              & $0.067 \pm 0.002$                                   & $0.183 \pm 0.053$                      & $10.4 s$                \\  
                          & LR                                                    & $0.924 \pm 0.048$                              & $0.351 \pm 0.219$                                   & $0.010 \pm 0.006$                      & $5.3 s$                 \\ \hline
\multirow{3}{*}{doc2vec}  & SVM                                                   & $0.936 \pm 0.060$                              & $0.036 \pm 0.020$                                   & $0.104 \pm 0.095$                      & $0.1 s$                 \\ 
                          & Random F.                                             & $0.678 \pm 0.126$                              & $0.099 \pm 0.010$                                   & $0.279 \pm 0.062$                      & $2.5 s$                 \\ 
                          & LR                                                    & $0.948 \pm 0.053$                              & $0.234 \pm 0.219$                                   & $0.007 \pm 0.006$                      & $0.3 s$                 \\ \hline
\end{tabular}
\end{adjustbox}

\end{table}

\color{black}

As for conventional pipelines, it is worth mentioning that the Random Forest classifier had a much lower performance for continuous textual representations (doc2vec and word2vec), which was expected due to the tree learning mechanism. It is noticed that its performance on sparse representations were very similar to the other algorithms, as they would naturally be easy to be divided by a composition of axis-parallel decision boundaries. SVM and LR had very similar performances in both continuous and sparse representations, but if we take F1-score as the first criterion, and Cross-Entropy and Brier Score as a tiebreaker, Logistic Regression wins in all conventional pipelines, but word2vec.

\textbf{The excellent performance of pipelines that use sparse features} is justified by a critical detail of the nature of the problem: there are words (BoW) and expressions (BoNG) that are a strong indication of whether the current subject is being interrupted or not. \textbf{These act as triggers, leading the classifier directly to the decision}. This is why weights, such as the commonly used TF-IDF, are not justified for this problem.
%In the examples of speeches made by the chairperson \emph{"Com a palavra a deputada Ana Mesquita para fazer um pedido de esclarecimento"} (in english: "Congresswoman Ana Mesquita is recognized for asking for a clarification") and \emph{"Tem a palavra o deputado André Ventura para responder à este bloco de perguntas"} (in english: "Congressman André Ventura is recognized to answer this block of questions") it is clear that expressions "answer this block of questions" and "asking for a clarification" indicate continuity. There are cases where expressions like these do not appear, but there are no signs of interruption either. Examples would be \emph{"queira terminar, deputada"} (in english: "please fiEnglishongresswomen"), \emph{"o seu tempo terminou sr. deputado”} (in english: ``gentleman, you are running out of time") or \emph{``Peço o silêncio do plenário para que o orador possa prosseguir”} (in english: ``I ask for silence from the chamber so the debater can proceed"). Yet, when there is an interruption, this is very clear in the chairperson's speech, for example in \emph{"Passo a palavra para o deputado Alberto Fonseca do PSD, para proferir uma nova declaração política"} (in English: "I give the floor for a new political statement from Congressman Alberto Fonseca from the PSD party"). 
It should be noted that \textbf{all models} (BERTimbau, SVM, LR, and Random Forest) \textbf{were able to deal with the class imbalance problem}. At the data level, Stratified K-fold CV was applied, and at the algorithm level, the F1-score metric was used to rank the models, in addition to the mechanisms that the models have for this: the complexity of BERT (with 334M parameters), balanced sub-sample from Random Forest and balanced class weighting from LR.

Another critical issue is that \textbf{BERTimbau handled the European Portuguese dialect well}. It is known that the model is pre-trained in the database brWaC~\cite{souza2020bertimbau}, which is mainly composed of data from the Brazilian Portuguese dialect. However, as the language is the same and the fine-tuning was performed with the transformer layers unfrozen, it was possible to adapt the pre-training for the specific domain (legislative debates) and the European dialect. On the other hand, it is remarkable that the BERT fine-tuning task can be costly and unnecessary, depending on the problem. Table \ref{tab:results} shows that the BERTimbau fine-tuning time is much higher than the training time of the other models. BERTimbau also makes greater use of memory space during fine-tuning and runtime. The execution time is known to be longer as well. Thus, \textbf{it is possible to say that we achieved the same results as the state-of-the-art in text classification with less computational effort}. Therefore, it is also recommended to consider traditional pipelines for tasks of the same type.

Finally, the results indicate that the \textbf{best classification pipeline for \texttt{DEBACER}} in the case of the Portuguese Parliament is the one that uses \textbf{Bag-of-N-Grams as text representation and Logistic Regression as classifier}. This is based not only on the performance metrics it obtained but mainly on being \textbf{the best trade-off between performance and computational resources spent} (time and memory in training and execution). Furthermore, in any pipeline configuration, the performance metrics results were good, which \textbf{demonstrates the viability of \texttt{DEBACER} as a solution to the proposed problem of partitioning moderated debates}.

%A case study shows the effectiveness of DEBACER using the user-defined keyword "corruption" and derived words (such as "corrupt", "corruptor", etc.). The graphics in Figure \ref{fig: graphics} show the number of direct and indirect citations of this word made by politicians (and their respective parties) in their political statements in the period from 2020/09/16 to 2021/02/25. The model used in this case study was created using \emph {Norm + BoNG + TSVD + SVM} configuration. The voter can verify this information, verifying the role of the politician and the parties on the topic of interest.

% The detecting topic interruption of DEBACER is used in our case study to determine the beginning and end of a new political statement; thus, we can see which citations of the word "corruption" were indirectly made by the congressperson involved in the same block of speech. With this information, we can generate a graphical report, such as the one in Figure \ref{fig: graphics} below,\textbf{which was created using \emph{Norm + BoNG + TSVD with SVM} configuration} which can easily provide information to an elector about the participation of their party or candidate in their topics of interest, making the monitoring of (anti)corruption ideas more clear and transparent by politicians.

\color{black}
% \begin{figure}[htb]
%     \centering
%     \fbox{\includegraphics[width=1\textwidth]{images/Racunho Tabela Resultados.png}}
%     \caption{Results}
%     \label{fig:Results}
% \end{figure}

% Please add the following required packages to your document preamble:
% \usepackage{multirow}
% Please add the following required packages to your document preamble:
% \usepackage{multirow}% Please add the following required packages to your document preamble:
% \usepackage{multirow}

\section{Conclusion and Future Work}
\label{sec:conclusion}

In this paper, we proposed \texttt{DEBACER}, a debate slicer method for partitioning speeches into blocks that share a common stage of a discussion. \texttt{DEBACER} groups speeches from different debaters that refer to the same subject. This method can be applied to moderated dialogues, where a moderator controls the session and passes the floor to whoever is the next to speak. The output blocks of our method allow the execution of other NLP tasks, such as topic modeling and sentiment analysis, as well as an assertive quantification of statistical information related to these data, such as citations to a specific topic or subject of interest, and its context.

There are many domains of \texttt{DEBACER}-able data. Among them, we can cite trials, public hearings, parliamentary sessions, and electoral debates, all sharing the structure of a moderated dialogue. For working properly in these domains, our algorithm's classification pipeline needs training within data from that domain, after which it will be ready to classify the moderator's speeches and partitioning the data into blocks.  In this work, \texttt{DEBACER} was validated on data from the minutes of the Portuguese Parliament.  We evaluated different pipelines to assess if the BERTimbau architecture, state-of-the-art in this task, is significantly better than more traditional ones. We show that a classic pipeline achieves scores statistically similar to BERTimbau, but with the advantage of having faster execution and training times and less memory usage. 

Some directions for future work will involve evaluating the performance of \texttt{DEBACER} and the techniques presented in this work for new datasets from different domains in order to validate their strength. In addition, we intend to explore applications where the proposed algorithm can be helpful as an intermediate step for NLP tasks in these datasets, such as topic modeling and opinion mining. Moreover, we want to investigate the potential and viability of employing cross-domain generalization strategies towards a universal classifier for \texttt{DEBACER}.

\section*{Acknowledgments}
This research was supported in part by \textit{Ita\'{u} Unibanco S.A.}, with the scholarship program of \textit{Programa de Bolsas Ita\'{u}} (PBI), and by the Coordenação de Aperfeiçoamento de Pessoal de Nível Superior (CAPES), Finance Code 001, and CNPQ (grant 310085/2020-9), Brazil.
Any opinions, findings, and conclusions expressed in this manuscript are those of the authors and do not necessarily reflect the views, official policy, or position of the Itaú-Unibanco, CAPES, and CNPq.

\bibliographystyle{sbc}
\bibliography{sbc-template}

\end{document}